\pdfoutput=1
\documentclass[11pt]{article}

\usepackage[]{acl}

\usepackage{times}
\usepackage{latexsym}

\usepackage[T1]{fontenc}
\usepackage[utf8]{inputenc}

\usepackage{microtype}

\usepackage{soul}
\usepackage{tablefootnote}
\usepackage[normalem]{ulem}
\usepackage{multirow}
\usepackage{makecell}
\usepackage{enumitem}
\usepackage{adjustbox}
\usepackage{amsmath}
\usepackage{array}
\usepackage{graphicx}
\usepackage{comment}
\usepackage{booktabs}
\usepackage{xcolor,colortbl} 
\usepackage{algorithm}

\usepackage{bbm}
\usepackage{tabularx}
\usepackage{hyperref}

\usepackage{graphicx}
\usepackage{subcaption}
\usepackage{arydshln}
\usepackage{array}
\usepackage{booktabs}
\usepackage{pifont}
\usepackage{cleveref}
\crefname{section}{§}{§§}
\usepackage{amsthm}
\usepackage{todonotes}
\usepackage{color}
\usepackage{amsmath, bm, amsfonts}

\newcommand{\modelname}{{\usefont{T1}{ppl}{m}{n}ProQA}}

\title{\modelname: Structural Prompt-based Pre-training for \\ Unified Question Answering}

\author{Wanjun Zhong$^{1}$\thanks{\ \ \ Indicates equal contribution}, Yifan Gao$^{4*}$, Ning Ding$^3$, Yujia Qin$^3$, Zhiyuan Liu$^3$, \\\textbf{Ming Zhou}$^5$, \textbf{Jiahai Wang$^1$, Jian Yin$^1$ and Nan Duan$^2$} \\
	$^1$ Sun Yat-sen University \quad $^2$ Microsoft Research Asia \\
    $^3$ Tsinghua University \quad $^4$ Chinese University of Hong Kong \quad $^5$ Langboat Technology  \\
	{\tt \{zhongwj25@mail2, wangjiah@mail,issjyin@mail\}.sysu.edu.cn}\\
	{\tt yfgao@cse.cuhk.edu.hk; \tt liuzy@tsinghua.edu.cn} \\
	{\tt \{dingn18, qyj20\}@mails.tsinghua.edu.cn} \\
	{\tt nanduan@microsoft.com}; \tt zhouming@chuangxin.com\\ 
}

\begin{document}
\maketitle
\begin{abstract}
Question Answering (QA) is a longstanding challenge in natural language processing.
Existing QA works mostly focus on specific question types, knowledge domains, or reasoning skills.
% The existing QA tasks are diverse in both question types, domains, and answer types, etc. 
The specialty in QA research hinders systems from modeling commonalities between tasks and generalization for wider applications.
% The speciality in QA research makes QA systems harder to model commonalities between tasks and hinder from wider application in a general scenario.
% However, primary studies mainly focus on a specific task, which makes the QA systems harder to model commonalities between tasks and avoids wider application in a general scenario.
To address this issue, we present \modelname, a unified QA paradigm that solves various tasks through a single model.
\modelname\ takes a unified \textit{structural prompt} as the bridge and improves the QA-centric ability by \textit{structural prompt-based pre-training}. 
% To address this issue, we present \modelname, a unified QA paradigm, which adopts a single model to solve various tasks, taking a unified \textit{structural prompt} as the bridge.
Through a structurally designed prompt-based input schema, \modelname\ concurrently models the knowledge generalization for all QA tasks while keeping the knowledge customization for every specific QA task.
% Structural prompt is capable of concurrently model the commonly required ability and difference between various tasks through a structurally designed prompt-based input schema. 
% Through pre-training with structural prompt-formulated large-scale synthesized corpus, \modelname\ empowers the model with general ability in both question answering and identifying the semantic meaning of the structural prompt. 
Furthermore, \modelname\ is pre-trained with structural prompt-formatted large-scale synthesized corpus, 
% which empowers the model with the general ability in both question answering and identifying the semantic meaning of the structural prompt.
which empowers the model with the commonly-required QA ability.
Experimental results on 11 QA benchmarks demonstrate that \modelname\ consistently boosts performance on both full data fine-tuning, few-shot learning, and zero-shot testing scenarios. Furthermore, \modelname\ exhibits strong ability in both continual learning and transfer learning by taking the advantages of the structural prompt.\footnote{The code is available at \url{https://github.com/zhongwanjun/ProQA}.}
% The code and trained checkpoints will be released to facilitate future research along this line.
% \footnote{We will release the code, pre-training corpus and pre-trained models upon acceptance.}

% Question Answering (QA) is a longstanding problem in natural language processing.
% Existing QA works mostly focus on a specific question type, knowledge domain, or reasoning skills.
% The speciality in QA research makes QA systems harder to model commonalities between tasks and avoids wider application in a general scenario.
% In this paper, we present \modelname, a single model to solve various QA tasks through the bridge of a unified \textit{structural prompt}.
% Through a structurally designed prompt-based input schema, \modelname\ concurrently models the knowledge generalization for all QA tasks while keep the knowledge customization for every specific task.
% Furthermore, \modelname\ is pre-trained with millions of structural prompt-formulated large-scale synthesized corpus, which empowers the model with general ability in both question answering and identifying the semantic meaning of the structural prompt. 
% Experimental results on 11 QA benchmarks demonstrate that \modelname\ consistently boosts performance on both full data fine-tuning, few-shot learning and zero-shot testing scenarios.
% \modelname\ also demonstrates strong ability in continual learning via its structual prompt.

\end{abstract}
\section{Introduction}
\begin{figure*}[t]
		\centering
		\includegraphics[width=0.9\textwidth]{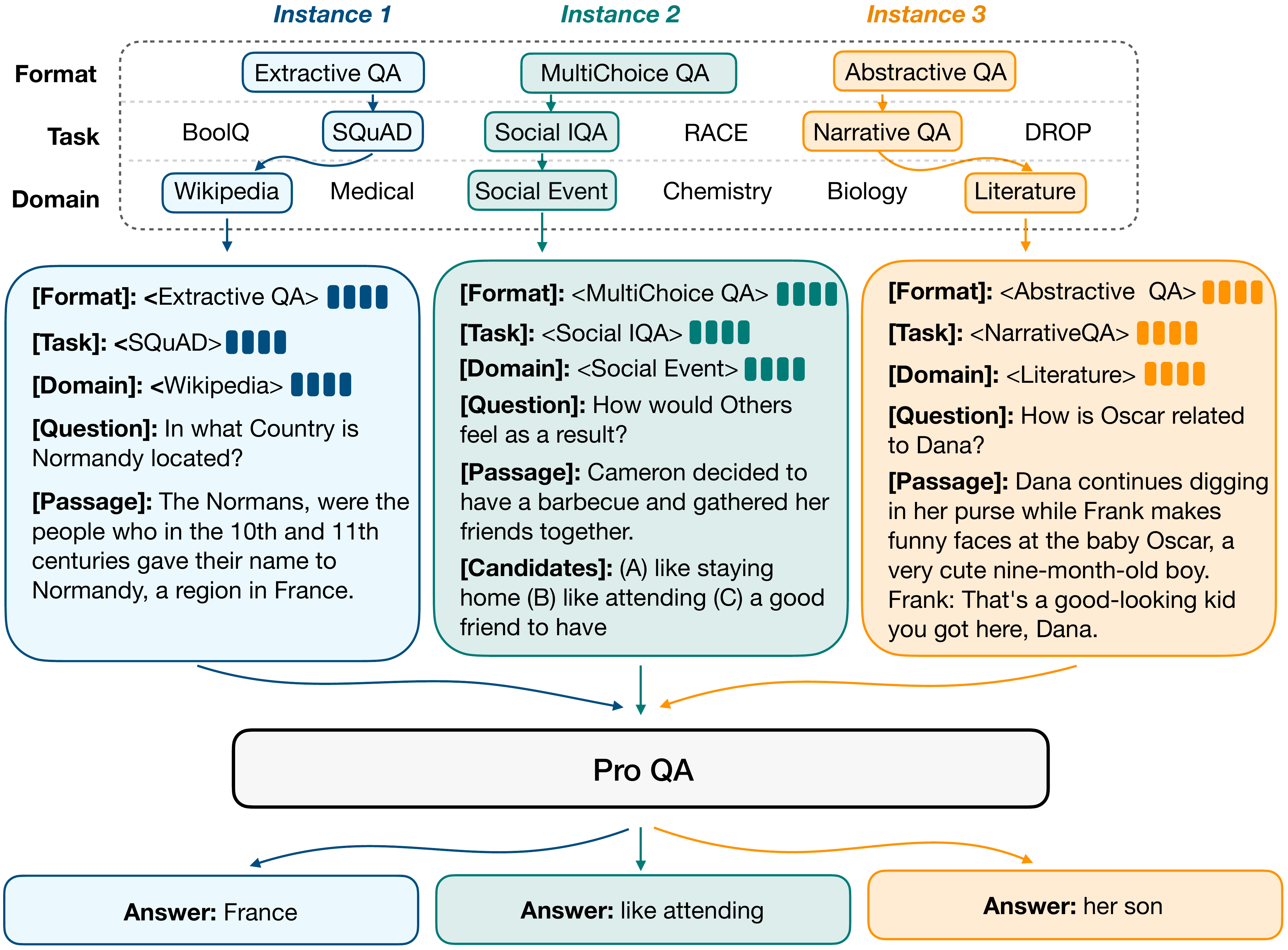}
		\caption{Approach overview of \modelname. Each box represents a specific instance formulated with the \textbf{structural prompt}, and \modelname\ is pre-trained with \textbf{structural prompt-based pre-training}. $[\quad]$ indicates \textit{special key indicator}, $<\quad>$ denotes \textit{hard prompt}, and colored squares denote \textit{continuous learnable soft prompts}.}
% 			We omit the fused evidence block from the figure for simplification.}
		\label{fig:proqa}
	\end{figure*}
	
%Version 4: modify 0111
Question Answering has long been an inspirational challenge in NLP research, and is viewed as the next-generation search engine and an essential tool for human beings to obtain knowledge \cite{etzioni2011search}. 
Many distinct datasets \cite{rajpurkar-etal-2016-squad,lai-etal-2017-race,kwiatkowski-etal-2019-natural,gao2021open} have been proposed along with the research trend on QA, involving very diverse question types (e.g., {\em extractive QA, abstractive QA, multiple-choice QA}), domains (e.g., {\em finance, daily events}), and answer types (e.g., {\em free-formed text, selected option}). 
The majority of previous works focus on tasks with specific question types \cite{lai-etal-2017-race,yang2018hotpotqa,gao2020discern} or specific domains \cite{trischler-etal-2017-newsqa,kwiatkowski-etal-2019-natural}. 
Recent research on large pre-trained language models \cite{brown2020language,bommasani2021opportunities} indicates that there may be tight connections among various tasks, which sheds light on a unified paradigm that can be potentially applied to solve various QA tasks to model their commonality. 

This observation motivates us to develop a unified QA model, which can model both the commonly-required QA ability and the difference between various QA tasks within a same paradigm. 
To achieve this goal, there are several key challenges needed to be addressed: 
(1) How to model commonalities and enhance transferability among different QA tasks in various domains/formats while reducing the conflict between them? 
(2) How to construct large-scale QA corpus as the high-quality QA-centric data is scarce for pre-training?

In light of this, we conceive \modelname, a unified QA paradigm, which builds up a general model to solve different QA tasks utilizing a \textbf{structural prompt} and improves commonly-required QA ability via \textbf{structural prompt-based pre-training}.

Firstly, to model the commonalities and distinguish task differences, we adopt a \textbf{structural prompt} to organize the inputs with a unified structurally designed input schema.
As illustrated in Fig.~\ref{fig:proqa}, given the complex components (e.g., {\em ``Domain"}, {\em ``Format"}, {\em ``Task"}, {\em ``Question"}, {\em ``Passage"}) as inputs, \modelname\ divides components into multiple key-value pairs, in which a specific component like {\em ``Question"} denotes a key, and the specific instance in this component is taken as the value. 
In this way, the model can discriminate different input components by key indicators and model the speciality of each task via task-specific values (learnable prompts).
% In this way, the model is capable of discriminating the semantics of different components with the component key indicator, and can better model the distinct between various task utilizing the task-specific value (learnable soft prompts) under the ``{\em Task}" key.

Secondly, to alleviate data sparsity problem and empower the model with transferability to the adaptation of new tasks, we conduct \textbf{structural prompt-based pre-training}. 
We first build a large-scale synthetic QA corpus automatically from Wikipedia, utilizing only a few seed datasets as the prior supervisions for pre-training corpus construction and finally covering primary QA formats. 
Then we format the pre-training data with the structural prompt, and teach the model to learn the general purpose QA-centric ability and the functionality of each component in the structural prompt via pre-training.
% We format the pre-training data with the structural prompt, and teach the model to learn the general purpose QA-centric ability as well as the semantic meaning of the key-value pairs in the structural prompt via pre-training. 
% Moreover, to mitigate the data sparsity problem, we automatically synthesize the pre-training corpus involving primary QA format types based on a large-scale unlabeled corpus Wikipedia, using only few seed datasets as the prior supervisions for the question-answer generation. 
% Thirdly, to mitigate the data sparsity issue in pre-training, we prepare a large-scale synthetic QA corpus from Wikipedia.
% The pre-training QA corpus starts from only a few seed datasets as the prior supervisions and finally covers primary QA formats.

We evaluate the effectiveness of \modelname\ on 11 downstream QA benchmarks, and the results show that our system achieves consistent performance boost in full data fine-tuning, few-shot learning, and zero-shot learning settings.
Experiments demonstrate that \modelname\ can better mitigate the catastrophic forgetting issue during continual learning by restoring the task-specific soft prompts residing in the structural prompt. 
Further analyses illustrate that our model has better transferability as it can be more quickly adapted to a newly involved task.
Ablation studies verify the effectiveness of both the soft prompt and prompt-based pre-training.

The contributions are summarized as follows:
\begin{itemize}[leftmargin = 15pt,topsep=0pt,noitemsep]
    \item We propose \modelname, a unified QA framework for solving various tasks within a single paradigm, taking an extensible and learnable structural prompt as the bridge.
    \item We enhance general QA-centric capabilities via structural prompt-based pre-training.
    \item Comprehensive experiments show that our model consistently improves the performance on 11 QA tasks especially in low-resource settings and exhibits better effectiveness in continual learning and few-shot transfer learning.
\end{itemize}

\section{Related Work}
\paragraph{Unifying QA formats.}
Despite vast diversity of current QA tasks in question type, answer type, answer source, and data domain~\citep{zeng2020survey}, there have been efforts in exploring a \textit{unified} format for various QA tasks. 
Some pioneered to demonstrate the generalization and transferability among different QA tasks~\citep{talmor-berant-2019-multiqa,dua2019orb,fisch2019mrqa}. Another line of works investigate multi-task learning for QA~\citep{mccann2018natural,shen2019multi,deng2019multi} by jointly training a single encoder to promote knowledge sharing. However, these methods typically require deploying distinct prediction heads for different tasks, which lead to poor scalability and flexibility when confronted with emerging QA tasks of new types.

To this end, inspired by the success of casting multiple tasks into the same text-to-text format~\citep{lewis-etal-2020-bart,raffel2019exploring}, researchers propose to learn a single model to unify various QA formats, alleviating the labor of task-specific designs~\citep{khashabi-etal-2020-unifiedqa,tafjord2021general}. However, these models (1) do not explicitly model the task or component characteristics, thus failing to properly disentangle the difference among QA tasks; and (2) overly rely on supervised data from specific tasks, which may not be available under data-scarce scenarios.

\paragraph{QA-centric pre-training.} Numerous efforts have been spent on improving PLMs' reasoning abilities with an intermediate pre-training stage before fine-tuning on target QA tasks, including (1) language modeling adaptation with salient span masking, which trains PLMs to recover randomly chosen~\citep{guu2020realm,wang2021can} or machine-generated~\citep{kang2020neural} masked named entities in the raw corpus; (2) training data augmentation \cite{zhong2022reasoning} with synthetic question-answer-context triples, such as generating (a) pseudo questions through 
% dual learning~\citep{tang2017question}, 
adversarial training~\citep{hosking-riedel-2019-evaluating,li2019improving}, knowledge bases~\citep{hu2021relation} or machine translation~\citep{lewis2019unsupervised,NEURIPS2020_7a677bb4}, (b) pseudo answers exploiting recurring spans~\citep{ram2021few} or rules based on heuristics~\citep{bian2021bridging} and (c) pseudo contexts via information retrieval~\citep{glass2019span}. Nevertheless, these works largely target at improving a certain reasoning ability for PLMs, and thus cannot be easily generalized to other QA tasks.

\paragraph{Prompts for PLMs.}
To effectively stimulate the knowledge acquired through pre-training, prompt-oriented fine-tuning is receiving increasing attention~\cite{liu2021pre,ding2021openprompt}, which re-formulates the objective of downstream tasks similar to that of pre-training by inserting manually designed~\citep{schick-schutze-2021-exploiting,schick-schutze-2021-just} or automatically searched~\citep{jiang-etal-2020-know,shin-etal-2020-autoprompt} hard \textit{prompt tokens} into the input text. 
Considering that discrete prompts may not be an optimal solution in the continuous embedding space, recent works ~\citep{li-liang-2021-prefix,hambardzumyan2021warp} proposed tunable soft prompts. It achieves satisfying performance especially when the model size grows extremely large~\citep{lester2021power}. Compared with the cumbersome parameters in PLMs, soft prompts are lightweight and pluggable, which paves the way for our goal of flexible adaptation to a new QA task.

\section{ProQA}
In this section, we detailedly describe the whole framework of \modelname\ for general purpose QA, which solves various QA tasks within the same paradigm.
\subsection{Overview}
\label{sec:overview}
\begin{figure}[!t]
		\centering
		\includegraphics[width=.4\textwidth]{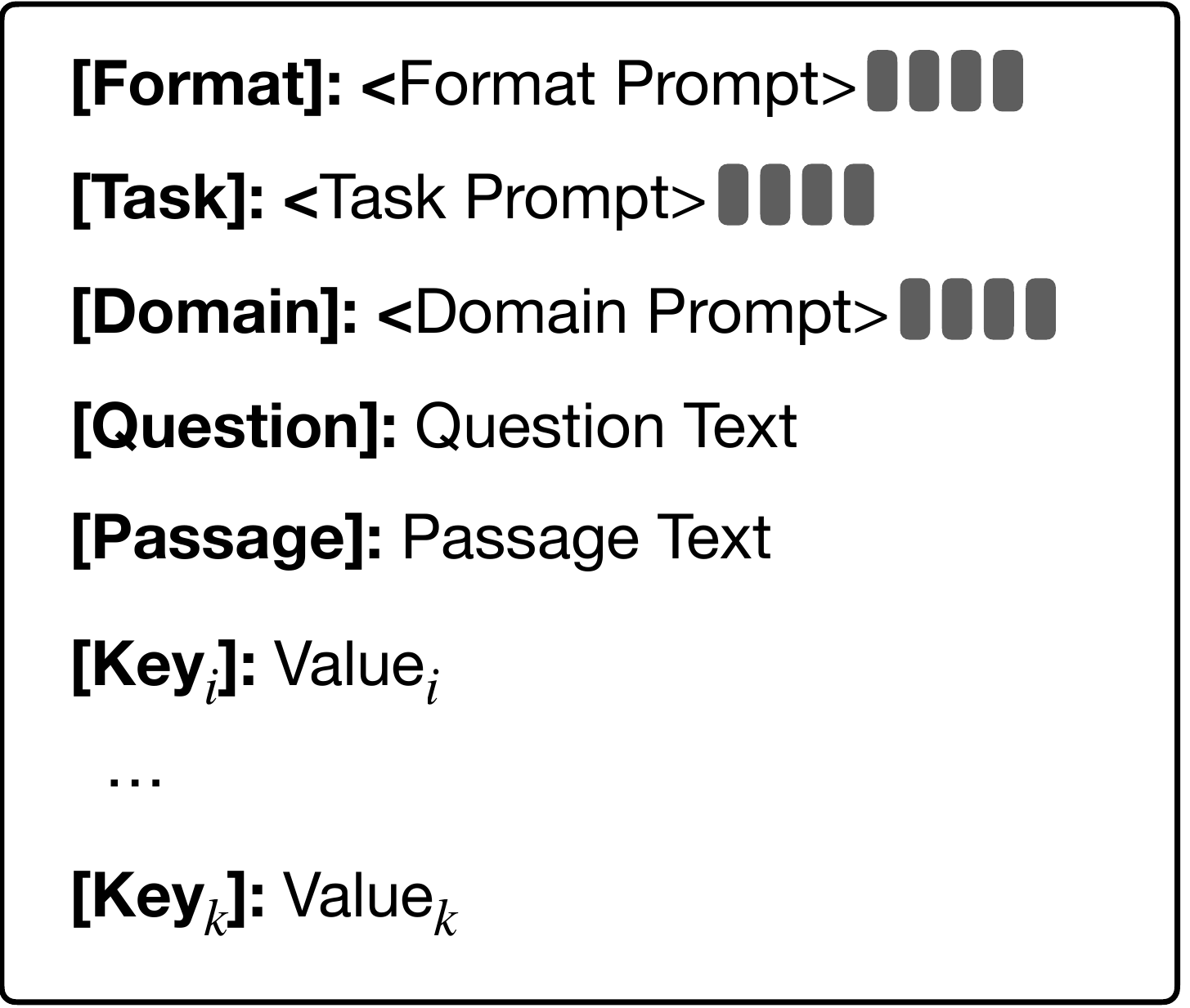}
		\caption{An illustration of the structural prompt. $[\quad]$ indicates special key indicator, $<\quad>$ denotes hard prompt, and grey squares indicate continuous soft prompts. }
% 			We omit the fused evidence block from the figure for simplification.}
		\label{fig:structural-prompt}
\end{figure}

As shown in Fig.~\ref{fig:structural-prompt}, we organize the inputs of various QA tasks with a unified \textbf{structural prompt} (\cref{sec:structural-prompt}), and adopt a unified model for question answering.
Then, to enhance the model in learning the QA-centric ability and the semantics of the structural prompt, we conduct \textbf{structural prompt-based pre-training} with synthetic pre-training corpus formatted with the structural prompt (\cref{sec:pre-training}). 

Inspired by \citet{khashabi-etal-2020-unifiedqa} and T5~\cite{raffel2019exploring}, we solve all downstream QA tasks with a unified text-to-text model. In this work, we mainly adopt T5 as the model backbone. 
Taking the structural prompt-based model input, the unified model generates the answer of the question.

\subsection{Structural Prompt}\label{sec:structural-prompt}
Here we detailedly illustrate the design of the structural prompt and its formatted input to the model.
\paragraph{Definition.} We organize complex QA task inputs with the structural prompt. 
As shown in Fig.~\ref{fig:structural-prompt}, the structural prompt consists of multiple $\{key: value\}$ pairs, 
where the $key$ represents a specific component\footnote{It is worth noting that ``{\em Format}" key denotes the format type (e.g., ``{\em MultiChoice QA}") of the task while the ``{\em Task}" key denotes a specific dataset (e.g., ``{\em SQuAD}").} (e.g., ``{\em Task}", ``{\em Format}", ``{\em Question}", etc.), 
and the $value$ has two possible types: (1) textual content (e.g., \textit{question}, \textit{passage}, \textit{options}) of the data instance; (2) task attributes (e.g., \textit{format, domain}) represented as the combination of a discrete \textit{hard prompt} and continuous \textit{soft prompts}.
The hard prompt is a predefined discrete description (we adopt a special token here), and the soft prompts are lightweight learnable and pluggable continuous embeddings that are proven to be parameter-effective in task adaptation~\cite{lester2021power}. 
The structural prompt-formatted examples are illustrated in Fig.~\ref{fig:proqa}.
In the case of the SQuAD dataset, ``$\langle$Format Prompt$\rangle$'', ``$\langle$Task Prompt$\rangle$'', ``$\langle$Domain Prompt$\rangle$'' will be ``$\langle$Extractive QA$\rangle$'', ``$\langle$SQuAD$\rangle$'', ``$\langle$Wikipedia$\rangle$'', respectively.
% \{``{\em Format}": {\em multiple-choice QA}, ``{\em Task}": {\em RACE}\}. 

To enhance the model in discriminating the functional difference between components, we adopt a \textbf{special key indicator} with learnable representation to represent each key.
Furthermore, to model the difference between several tasks/domains/formats, we also adopt learnable and storable \textbf{specific soft prompts} as the $value$ to represent their customized characteristics, which makes the model more flexible for task adaptation.
% Moreover, since the semantics of task and format are hard to be modeled by a single hard prompt (discrete language phrase), we also adopt several task/format-specific soft prompt tokens as the value of the domain ``{\em Task/Format}".
% Soft prompts are trainable continuous embeddings that are proven to be parameter-effective in task adaptation~\cite{lester2021power}. 

As a result, the structural prompt can empower the model in the following aspects:
(1) modeling knowledge generalization of various tasks utilizing a unified input schema;
(2) discriminating different components with the special $key$ indicator; 
(3) customizing the speciality of each task/format/domain with learnable and storable soft prompts as the $value$ under corresponding keys.
% the semantics of input components are differentiated by the variance of the $domain$ key.
% Besides, since the task/format-specific soft prompts are trainable and storable tiny vectors, which can record the characteristic for each task/format, and thus can help the model to distinguish between various tasks. 
% It can also be potentially utilized for continual learning by restoring task-specific prompts for recalling learned tasks.
\paragraph{Input Representation.}
Specifically, given a structural prompt-formatted instance, we describe the specific representation of the model input. 
We firstly translate $k^{th}$ key to a {\em key indicator $D_k$} (a special token), which is attached by the tokens $V_k$ of the specific value to form a token sequence. It is further represented as $\bm{E}_k=Embedding([D_k; V_k])$. The representation of $D_k$ is initialized and updated during training.
% Then we concatenate all the domain-value representation $X_k$ to form 
Since we use soft prompts $\bm{P}_{\text{task}}$/$\bm{P}_{\text{format}}$/$\bm{P}_{\text{domain}}$ as the value of the corresponding key and they are commonly required for all the tasks, we prepend them to the input for convenience and concatenate all the $\bm{E}_k$ to form the final model input $\bm{X}$:
\begin{equation}
\bm{X} = [\bm{P}_{\mathrm{domain}}; \bm{P}_{\mathrm{format}}; \bm{P}_{\mathrm{task}}; \bm{E}_{1}; ...; \bm{E}_{k}]
\end{equation}
It is also worth noting that the representations $\bm{D}$ of key indicators and the soft prompts $\bm{P}$ are jointly trained with the main model parameters during pre-training for learning the semantics of the structural prompt. 
Moreover, after being tuned by various tasks, the soft prompts $\bm{P}$ can be stored to record the customized task-specific characteristics.
% \footnote{We mainly use the format/task-specific soft prompts in our experiments.}.

\subsection{Structural Prompt-based Pre-training}
\label{sec:pre-training}
In this part, we introduce how we conduct structural prompt-based pre-training to help the model in learning commonly-required QA ability and the semantics of the structural prompt during pre-training to facilitate the adaption of the structural prompt to downstream tasks. 

\paragraph{Task Formulation.}
Along with the structural prompt-based paradigm, we manifest various exemplary QA format types (i.e., {\em Extractive QA}, {\em Abstractive QA}, {\em Multiple-choice QA} and {\em Yes/No QA}) for pre-training to inject the general QA-centric ability.
% Given the multi-format QA pre-training corpus, 
% \modelname\ process all formats 
Given the multi-format QA pre-training corpus, we transform all QA formats according to the proposed structural prompt, which enables joint pre-training while keeping the differences among various formats. 
% all formats go through the transformation according to the proposed structural prompt, which enables joint pre-training while keeping the difference between various formats.
% we first formulate the data belongs to different QA format types with the proposed structural prompt, which results in a unified pre-training corpus while keeping the difference between various formats. 
Taking a structural prompt-formatted instance as the input and a free-form answer as the output, the task is further tailored to a QA task with the encoder-decoder model. 

\paragraph{Pre-training Corpus Construction.}
% Since the data sparsity problem is extremely severe for building the QA-centric pre-training corpus as it is impractical and laborious to obtain the large-scale high-quality annotated data, and is also hard to generate QA-centric self-supervised data using rule-based methods (e.g., token masking or sentence reordering).

When we prepare the QA pre-training corpus, data sparsity problem is extremely severe because 
(1) it is impractical and laborious to obtain a large-scale high-quality annotated data for pre-training
 and (2) it is hard to generate QA-centric self-supervised data using rule-based methods (e.g., token masking or sentence reordering).
In this work, inspired by \citet{lewis-etal-2021-paq}, we adopt a \textbf{generation-filtering based corpus construction method} to synthesize a large-scale pre-training corpus, based on a large-scale unlabeled Wikipedia corpus with almost 6 million passages. 

Typically, the general generation-filtering process consists of the following components:
\begin{itemize}[leftmargin = 15pt,topsep=0pt,noitemsep]
    \item[1.] A QA-pair generation model $g_{qa}(q,a|c)$: Given a passage $c$ as input, $g_{qa}(q,a|c)$ generates $q\ \texttt{[SEP]}\ a$ as the output sequence including a pair of question $q$ and its answer $a$.
    \item[2.] A filtering QA language model $f(a|q,c)$ for filtering the generated QA-pairs to ensure the quality and consistency of the question and the answer. 
    $f(a|q,c)$ is a conditional-probability-based approach to filter out QA pairs softly. It scores a QA pair $(q,a)$ with the likelihood of the answer $a$ conditioned on the passage $c$ and question $q$.
    The QA-pairs with scores higher than a threshold will be kept for pre-training.
\end{itemize}
We adopt the same text-to-text pre-trained model T5 described in \cref{sec:overview} as the model backbone of both the generation and filtering model.

To ensure the reliability of the generation and filtering models, we inevitably select a few seed datasets (typically one for each QA format type) as the prior supervisions to train these models.
It is worth mentioning that, we avoid using more supervised data for corpus construction, because we expect the whole paradigm to have better expandability. 
In other words, if we want to extend the paradigm for a newly-involved QA format type but with limited supervised data, we can utilize these data to automatically create a synthetic large-scale pre-training corpus.

More specifically, the construction method has little variance for different formats according to their input components. 
For \textbf{\em Extractive QA} and \textbf{\em Abstractive QA}, we adopt the aforementioned general method to synthesize QA-pairs. 
We also tried to first extract answers using rule-based method (extracted named-entities or key phrases), and only generate questions.
We empirically find that this method performs much worse as it involves simple bias of the rule-based method.
As the inputs for \textbf{\em Multiple-Choice QA} involve a new component ``{\em Candidate Answers}", we adopt a distractor (negative options) generation model $g_{neg}(o|c,q,a)$ to generate three negative options $o$. 
For \textbf{\em Yes/No QA}, we simply generate questions by taking {\em True/False} as the corresponding answers.
Further details are described in Appendix \ref{sec:app_implementation}.

\section{Experimental Setup}

% Table generated by Excel2LaTeX from sheet 'latex'
\begin{table}[!t]
  \centering
  \resizebox{1.0\columnwidth}{!}{
    % Table generated by Excel2LaTeX from sheet 'latex'
    \begin{tabular}{ccccc}
    \toprule
    Format & Dataset & \#Train & \#Dev & QA Skills \\
    \midrule
    \multirow{3}[2]{*}{Extractive QA} & $\text{SQuAD}^{*}$ & 87k   & 10k   & Word Matching \\
          & Quoref & 22k   & 2k    & Coreference Reasoning \\
          & NewsQA & 76k   & 4k    & Word Matching \\
    \midrule
    \multirow{3}[2]{*}{Abstractive QA} & $\text{NarQA}^{*}$ & 65k   & 21k   & Story Understanding \\
          & DROP  & 77k   & 9k    & Discrete Reasoning \\
          & NQOpen & 79k   & 3.6k  & Multi-passage Understanding \\
    \midrule
    \multirow{5}[2]{*}{MultiChoice QA} & $\text{RACE}^{*}$  & 87k   & 4k    & Multi-sentence Reasoning \\
          & DREAM & 6k    & 2k    & Dialog Reasoning \\
          & MCTest & 1.4k  & 320   & Multi-sentence Reasoning \\
          & OBQA  & 4k    & 501   & Common Knowledge \\
          & SIQA  & 33.4k & 2.2k  & Commonsense Reasoning \\
    \bottomrule
    \end{tabular}%
    }
    \caption{Dataset statistics and required language understanding skills. Datasets with * denote seed datasets for preparing pretraining data.}
  \label{tab:dataset}%
\end{table}%

% Table generated by Excel2LaTeX from sheet 'latex'
\begin{table*}[!t]
  \centering
  \resizebox{1.0\textwidth}{!}{
    % Table generated by Excel2LaTeX from sheet 'latex'
    \begin{tabular}{cl|ccc|ccc|ccccc|c}
    \toprule
    \multirow{2}[4]{*}{Setting} & \multicolumn{1}{c|}{\multirow{2}[4]{*}{Dataset}} & \multicolumn{3}{c|}{ExtractiveQA} & \multicolumn{3}{c|}{AbstractiveQA } & \multicolumn{5}{c|}{MultiChoiceQA}    & \multirow{2}[4]{*}{Avg} \\
\cmidrule{3-13}          &       & SQuAD & Quoref & NewsQA & NarQA & DROP  & NQOpen & RACE  & DREAM & MCTest & OBQA  & SIQA  &  \\
    \midrule
    \multirow{4}[2]{*}{Full-Data} & T5    & 83.4  & 64.9  & 45.2  & 49.3  & 45.0  & 42.3  & 67.9  & 54.8  & 44.4  & 49.6  & 64.1  & 55.5 \\
          & UnifiedQA & 84.4  & 74.8  & 45.3  & 49.6  & 45.1  & 42.5  & 71.6  & 67.6  & 83.1  & 57.6  & 64.9  & 62.4 \\
          & \modelname\ (qapair) & 84.9  & 76.6  & \textbf{50.8}  & 49.8  & \textbf{55.0} & 43.2  & \textbf{73.6} & 72.9  & 85.0  & \textbf{61.6} & \textbf{67.5} & \textbf{65.5} \\
          & \modelname\ (paq) & \textbf{85.3} & \textbf{76.8} & {50.4} & \textbf{50.1} & 52.5  & \textbf{43.9} & 73.2  & \textbf{73.3} & \textbf{85.9} & 61.4  & 67.2  & \textbf{65.5} \\
    \midrule
    \multirow{4}[2]{*}{Few-Shot} & T5    & 6.7   & 14.6  & 20.5  & 3.4   & 5.8   & 11.9  & 26.2  & 34.7  & 38.1  & 29.0  & 32.4  & 20.3 \\
          & UnifiedQA & \underline{82.0}  & 38.2  & 34.2  & \underline{49.1}  & 22.2  & 31.6  & \underline{53.0}  & 57.4  & 73.8  & 41.2  & 42.8  & 48.1 \\
          & \modelname\ (qapair) & \underline{82.9}  & 44.2  & 41.1  & \underline{49.1}  & 24.9  & 33.3  & \underline{63.4}  & 64.5  & 82.5  & \textbf{46.2} & 49.1  & 52.8 \\
          & \modelname\ (paq) & \underline{\textbf{84.4}} & \textbf{52.2} & \textbf{42.1} & \underline{\textbf{49.2}}  & \textbf{27.1} & \underline{\textbf{36.0}} & \underline{\textbf{66.5}} & \textbf{66.0} & \textbf{84.1} & 44.8  & \textbf{49.4} & \textbf{54.7} \\
    \midrule
    \multirow{4}[2]{*}{Zero-Shot} & T5    & 0.0   & 0.0   & 0.0   & 3.5   & 2.0   & 1.5   & 24.1  & 34.2  & 27.5  & 21.9  & 33.2  & 13.5 \\
          & UnifiedQA & \underline{80.7}  & 27.9  & 31.4  & \underline{48.3}  & 18.0  & 30.9  & \underline{53.0}  & 57.0  & 73.4  & 35.9  & 40.3  & 45.2 \\
          & \modelname\ (qapair) & \underline{80.4}  & 30.5  & 30.7  & \underline{48.1}  & 17.0  & 33.0  & \underline{62.6}  & 64.3  & \textbf{81.3} & 36.0  & \textbf{47.2} & 48.3 \\
          & \modelname\ (paq) & \underline{\textbf{81.3}} & \textbf{42.1} & \textbf{31.8} & \underline{\textbf{48.4}} & \textbf{19.7} & \underline{\textbf{36.0}} & \underline{\textbf{65.9}} & \textbf{65.2} & \textbf{81.3} & \textbf{38.6} & 46.7  & \textbf{50.6} \\
    \bottomrule
    \end{tabular}%
    }
\caption{Main results on 11 downstream QA datasets under full-data fine-tuning, few-show learning, and zero-shot learning settings. 
Since the supervisions of seeds datasets are used in the pre-training corpus construction which may introduce bias in few-shot and zero-shot settings, results on these corresponding entries are underlined.}
\label{tab:main}%
% \vspace{-1em}
\end{table*}%

\subsection{Datasets and Evaluation Metrics}
% 1. list datasets
% 2. introduce the underlying understanding skills required from each dataset
% 3. evaluation metrics for each datasets
% 4. other dataset-specific details
We consider three formats of QA datasets in our experiments: Extractive QA, Abstractive QA and Multiple-Choice QA\footnote{We also include Yes/No QA in our pilot study. We do not consider it in our main experiments because datasets in this format are extremely rare. Results on this QA formats are shown in Appendix~\ref{sec:app_results_yesno}.}.
For each QA format, we select one \textit{seed} dataset for preparing the large-scale pre-training data.
The seed dataset is used to train the question-answer generation and filtering models in the process of pre-training corpus construction.
In total, the experiments are conducted on 11 QA datasets with three different formats and various language understanding abilities.
An overview of datasets used in the experiments and their required QA skills are summarized in Table~\ref{tab:dataset}.

\paragraph{Extractive QA.}
We take SQuAD 1.1~\cite{rajpurkar-etal-2016-squad} as the seed dataset for extractive style QA.
In addition, we consider NewsQA~\cite{trischler-etal-2017-newsqa} and Quoref~\cite{dasigi-etal-2019-quoref} to evaluate the generalization ability of models.
The EM (Exact Match) score between the extracted span and the gold answer span is used as the evaluation metric for extractive QA.

\paragraph{Abstractive QA.}
Narrative QA (NarQA) \cite{kocisky-etal-2018-narrativeqa} is taken as the seed dataset for Abstractive QA.
DROP~\cite{dua-etal-2019-drop} and the open-domain version of NaturalQuestions (NQOpen)~\cite{kwiatkowski-etal-2019-natural} are also considered.
Passages for each question in NQOpen are retrieved by the dense passage retriever~\cite{karpukhin-etal-2020-dense} and are concatenated into a sequence.
We use ROUGE-L~\cite{lin-2004-rouge} metric for NarQA and F1 score for DROP and NQOpen.

\paragraph{Multiple-Choice QA.}
For multiple choice QA, the following datasets are considered: RACE~\cite{lai-etal-2017-race} (seed dataset), DREAM~\cite{sun-etal-2019-dream}, MCTest~\cite{richardson-etal-2013-mctest}, OpenBookQA (OBQA)~\cite{mihaylov-etal-2018-suit}, Social IQa (SIQA)~\cite{sap-etal-2019-social}.
OBQA does not have contexts (reading comprehension passages).
The context for DREAM is in the dialogue style and we concatenate them into a sequence as the passage input. 
We select the option with the highest textual similarity with the generated answer as the final answer.
We compute the accuracy of the correct options for all multiple choice QA datasets.

% Table generated by Excel2LaTeX from sheet 'latex'
\begin{table*}[!t]
  \centering
  \resizebox{1.0\textwidth}{!}{
    \begin{tabular}{lcccccccccc}
    \toprule
    \multicolumn{1}{c}{\multirow{2}[2]{*}{Methods}} & \multicolumn{10}{c}{Task A$\rightarrow$Task B} \\
          & EX$\rightarrow$EX & EX$\rightarrow$AB & EX$\rightarrow$MC & AB$\rightarrow$EX & AB$\rightarrow$AB & AB$\rightarrow$MC & MC$\rightarrow$EX & MC$\rightarrow$AB & MC$\rightarrow$MC & Avg \\
    \midrule
    Task B Model & 20.5\% & 26.2\% & 13.0\% & 8.9\% & 6.3\% & 6.5\% & 4.6\% & 4.9\% & 0.9\% & 9.9\% \\
    Task B Model (w/ Task A Prompt) & 17.1\% & 10.6\% & 6.7\% & 3.1\% & 3.1\% & 2.2\% & 0.4\% & 0.7\% & -0.5\% & 4.3\% \\
    \bottomrule
    \end{tabular}%
}
\caption{
Continual learning results for averaged performance drops compared with the original task A results under different task learning orders (\textbf{lower is better}). 
Negative number means the performance improves compared with the original task A results.
EX: Extractive QA; AB: Abstractive QA; MC: Multiple-chioce QA.}
\label{tab:continual}%
\end{table*}%

\subsection{Approaches}
% We compare our proposed structured prompt-based pre-trained model \modelname\ with T5 and UnifiedQA:

\paragraph{T5}~\cite{raffel2019exploring} is a unified text-to-text pre-training framework that covers all text-based language problems. 
We use \texttt{google}/\texttt{t5-v1\_1-base} from HuggingFace Transformers \cite{wolf-etal-2020-transformers} that is only pre-trained on C4 excluding any supervised training dataset (e.g., QA datasets).
    
\paragraph{UnifiedQA}~\cite{khashabi-etal-2020-unifiedqa} crosses the format boundaries of different QA tasks by formulating them into text-to-text tasks under T5.
It directly concatenates all inputs via \texttt{\textbackslash n} into a sequence and feeds it into T5 for predicting the answer.
We train our own UnifiedQA model on the combination of three aforementioned seed datasets, namely SQuAD, NarQA, and RACE.
    
\paragraph{{\usefont{T1}{ppl}{m}{n}\textbf{ProQA}}} is our proposed structural prompt-based pre-training approach.
\modelname\ is pre-trained jointly on three formats of pre-training corpus: Extractive QA, Abstractive QA, and Multiple-Choice QA.
This approach using corpus prepared from QA-pair generation-filtering model described in \cref{sec:pre-training} is named as \modelname\ (qapair).
Additionally, we leverage the off-the-shelf large-scale QA pairs from Probably-Asked Questions/PAQ \cite{lewis-etal-2021-paq}, and replace our extractive QA pre-training corpus by a subset of PAQ (abstractive QA and multiple-choice QA corpus remains unchanged).
PAQ provides a refined pipeline that introduces learned models on every step of QA pair generation, i.e., passage selection, answer identification, question generation, and filtering.
We name this variant as \modelname\ (paq).

For every downstream QA dataset, we start from the above pre-trained models and conduct experiments under full-data fine-tuning, few-shot learning, and zero-shot learning settings. 
For few-shot learning, we randomly sample 32 instances from the training set.

\section{Results and Analyses}

\subsection{Main Results}
Main results are shown in Table~\ref{tab:main}, and we have the following observations:
% 1. UnifiedQA & Pro QA > T5: knowledge generalization
% 2. Pro QA > UnifiedQA: knowledge customization
% 3. Compare paq and qapair
\begin{itemize}[leftmargin = 15pt,topsep=0pt,noitemsep]
    \item QA-centric pre-trained models, namely UnifiedQA and \modelname, outperform T5 by a large margin on both seed datasets and non-seed datasets.
    This is because there is some transferable knowledge across different QA tasks. Once the model is pre-trained by any QA task, the learned knowledge can be generalized to any other datasets.
    \item \modelname\ demonstrates better knowledge customization ability than UnifiedQA -- \modelname\ beats UnifiedQA by a large margin in few-shot and zero-shot settings.
    This is because (1) the hard and soft prompts in the structural prompt enable better knowledge customization for every QA task, especially the ``Task'' key-value pair that is different for every QA task; (2) structural prompt-based pretraining empowers \modelname\ to adapt faster (\cref{sec:convergence}) and better (Table~\ref{tab:main}) to these non-seed datasets.
    \item Comparing \modelname\ (qapair) and \modelname\ (paq), we find that \modelname\ (paq) performs better in most scenarios.
    Presumably, PAQ provides high quality pre-training corpus through its pipelined approach -- there are in total four BERT-sized models to be prepared for generating PAQ corpus. 
    Instead, our proposed QA pair generation approach is simple and can be applied to not only Extractive QA but also Abstractive QA and Multiple-choice QA in the pre-training corpus construction process.
\end{itemize}

\subsection{Continual Learning via Soft Prompt}

One benefit of introducing soft prompt in \modelname\ is that it can potentially mitigate the catastrophic forgetting issue when adapting to a new task.
If \modelname\ is \textbf{sequentially fine-tuned on task A and task B under few-shot setting}, it can load task A soft prompt back when it is evaluated again on the task A.
The plug-in flexibility of \modelname\ brings huge improvements compared with its counterpart that keeps the task B soft prompt.

We conduct continual learning  by setting task A and B as different combinations among datasets with formats\footnote{Note that we consider two tasks in continual learning because we also want to directly investigate the task adaptation to-and-fro the same format (e.g., MC $\rightarrow$ MC) or different formats (e.g., AB $\rightarrow$ EX).}: Extractive QA (EX), Abstractive QA (AB), and Multiple-choice QA (MC).
Formally, we first adapt \modelname\ to task A by few-shot learning to obtain the model A: $f_{\theta}^{A}$ with performance $s^A$. Then we sequentially adapt $f_{\theta}^{A}$ to task B and receive task B model $f_{\theta}^{AB}$. 
We evaluate performance of the model $f_{\theta}^{AB}$ on task A under two settings: (1) direct testing (task-B prompt) (2) first restoring the learned task-A prompt from $f_{\theta}^{A}$ to the model $f_{\theta}^{AB}$ and then testing. Performance of the two settings are denoted as $s^{AB}$ and $s^{AB'}$, respectively.
We evaluate the continual learning performance under these two settings with the \textbf{percentage of the performance drop} on the task A: 
``\textit{Task B Model}''= $\frac{s^A - s^{AB}}{s^A}$, and ``\textit{Task B Model (w/ Task A Prompt)}'' = $\frac{s^A - s^{AB'}}{s^A}$.

\iffalse
Formally, we firstly fine-tune the pre-trained \modelname\ $f_{\theta}$ on task A under few-shot learning setting. 
The fine-tuned model and the evaluation results on Task A are $f_{\theta}^{A}$ and $s^A$, respectively.
Starting from $f_{\theta}^{A}$, we continue to adapt it to task B and receive $f_{\theta}^{B}$.
We evaluate the fine-tuned task B model on the task A under two approaches: (1) ``Task B Model'': we directly evaluate the task B model $f_{\theta}^{B}$ on the Task A and denote the result as $s^B$; (2) ``Task B Model (w/ Task A Prompt)'': we replace the soft prompt in the task B model $f_{\theta}^{B}$ by the soft prompt in the previous task A model $f_{\theta}^{A}$, and evaluate accordingly. The evaluation score is denoted as $s^{B'}$.
The catastrophic forgetting degrees for these two approaches are measured by the performance drop on the task A before and after the fine-tuning on the task B: ``Task B Model''= $\frac{s^A - s^B}{s^A}$; ``Task B Model (w/ Task A Prompt)'' = $\frac{s^A - s^{B'}}{s^A}$.
\fi
As shown in Table~\ref{tab:continual}, the catastrophic forgetting issue does exist when evaluating task A with task B model (``\textit{Task B Model}'') directly.
The performance drops as large as 26.2\% for EX$\rightarrow$AB.
However, restoring task A prompt brings huge improvements across all task combinations (``\textit{Task B Model w/ Task A Prompt}'').
% After training on task B, 
It is surprising to see that restoring task A prompt could sometimes even improve task A performance (MC$\rightarrow$MC = $-$0.5\%).
Presumably, sequential learning two tasks under the same question format (MC) makes the model learn the transferable knowledge while restoring task A prompt brings task-specific knowledge.
Detailed experimental results on the 33 combinations of datasets can be found in Appendix~\ref{sec:app_results_continual}.

\begin{figure}[t!]
\centering
\includegraphics[width=1.0\columnwidth]{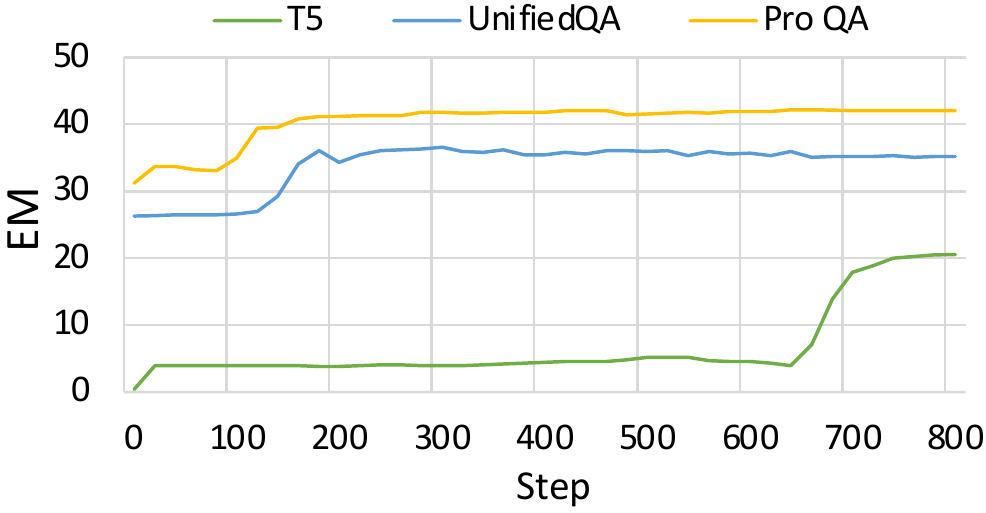}
\caption{
The few-shot learning curves of EM scores on the validation set of the NewsQA task.
}
% \vspace{-1em}
\label{fig:convergence}
\end{figure}

\subsection{Convergence Analysis}\label{sec:convergence}

We investigate the effectiveness of pre-training by compare the step-wise performance under few-shot learning setting.
The learning curves of EM scores on the validation set of the NewsQA task is shown in Figure~\ref{fig:convergence}.
Out of the three models, T5 convergences slowest because it does not have any QA-centric knowledge while our proposed \modelname\ adapts fastest and best.
Moreover, we find that UnifiedQA EM score rapidly saturates
and eventually degrades slightly, suggesting that the model overfits under the few-shot setting.
On the counterpart, our \modelname\ continues  to improve and never degrades because the hard and soft prompt inside the structural prompt balance the knowledge generalization and knowledge customization well.

% % Table generated by Excel2LaTeX from sheet 'latex'
% \begin{table}[!t]
%   \centering
%   \resizebox{1.0\columnwidth}{!}{
%     \begin{tabular}{lccc}
%     \toprule
%     Model & NewsQA & NQOpen & DREAM \\
%     \midrule
%     \modelname\ & 42.1  & 36.0  & 66.0 \\
%     \quad w/o soft prompt & 38.5  & 32.9  & 64.5 \\
%     \quad\quad w/o pretraining & 20.5  & 10.7  & 35.1 \\
%     UnifiedQA + Pre-train Corpus & 37.3  & 32.4  & 59.6 \\
%     \bottomrule
%     \end{tabular}%
%     }
%     \caption{Ablation study results on three non-seed datasets under different QA formats (extractive, abstractive, multiple-choice).
%     % (NewsQA$\rightarrow$Extractive; NQOpen$\rightarrow$Abstractive; DREAM$\rightarrow$MultiChoice).
%     }
%   \label{tab:ablation}%
% %   \vspace{-1em}
% \end{table}%

% Table generated by Excel2LaTeX from sheet 'latex'
\begin{table}[!t]
  \centering
  \resizebox{1.0\columnwidth}{!}{
    \begin{tabular}{clccc}
    \toprule
    Setting & \multicolumn{1}{c}{Model} & NewsQA & DROP  & DREAM \\
    \midrule
    \multirow{4}[1]{*}{Full-Data} & \modelname\ & 50.4  & 52.5  & 73.3 \\
          & \quad w/o soft prompt & 48.6  & 51.2  & 69.9 \\
          & \quad\quad  w/o pretraining & 48.1  & 44.5  & 68.4 \\
          & UnifiedQA + Pre-train Corpus & 46.8  & 50.3  & 69.4 \\
    \midrule
    \multirow{4}[1]{*}{Few-Shot} & \modelname\ & 42.1  & 27.1  & 66.0 \\
          & \quad w/o soft prompt & 38.5  & 24.3  & 64.5 \\
          & \quad\quad  w/o pretraining & 20.5  & 4.8   & 35.1 \\
          & UnifiedQA + Pre-train Corpus & 37.3  & 23.5  & 59.6 \\
    \midrule
    \multirow{4}[2]{*}{Zero-Shot} & \modelname\ & 31.8  & 19.7  & 65.2 \\
          & \quad w/o soft prompt & 29.8  & 19.0  & 63.5 \\
          & \quad\quad  w/o pretraining & 0.0   & 0.2   & 32.6 \\
          & UnifiedQA + Pre-train Corpus & 26.0  & 18.1  & 58.0 \\
    \bottomrule
    \end{tabular}%
    }
    \caption{Ablation study results on three non-seed datasets under different QA formats (extractive, abstractive, multiple-choice).
    % (NewsQA$\rightarrow$Extractive; NQOpen$\rightarrow$Abstractive; DREAM$\rightarrow$MultiChoice).
    }
  \label{tab:ablation}%
%   \vspace{-1em}
\end{table}%

\subsection{Ablation Study}
An ablation study is conducted to unveil the effectiveness of every component in \modelname.
We consider three variants of \modelname: 
(1) \modelname\ without the soft prompt in its structural prompt;
(2) \modelname\ further without prompt-based pre-training.
(3) UnifiedQA + Pre-train Corpus is the UnifiedQA model pre-trained on our prepared large-scale synthetic QA corpus.
Results on three non-seed datasets under different QA formats are shown in Table~\ref{tab:ablation}.
We find that removing the soft prompt from the model disables the task-specific knowledge learned during pre-training.
Moreover, removing the prompt-based pretraining drastically hurts the performance as the equivalent model (T5 + hard structural prompt) does not have any QA knowledge.
Finally, UnifiedQA + Pre-train Corpus could not compete with \modelname, showing that our proposed structural prompt earns better balance between knowledge generalization and knowledge customization than UnifiedQA.

\section{Discussion}
In this section, we discuss on how to extend the \modelname\ to a new task even with a new schema, and sheds light on potential future directions. 
% \begin{itemize}[leftmargin = 15pt,topsep=0pt,noitemsep]

    \textit{1)} \textit{Task Adaptation with Structural Prompt:}
    The design of structural prompt empowers \modelname\ with better expandability.
    In our main experiments, we adopt 3 format types and 11 QA tasks. 
    In the future, we can adapt \modelname\ to more tasks, formats, domains, and new input schema. 
    Intuitively, when being adapted to a new task with unseen format/domain, \modelname\ can initialize the specific soft prompts and learn the characteristic of the new domain/task through model training. 
    % In this way, the general knowledge reside in \modelname\ can also be adapted to the new task while learning its characteristic, because the model already know the semantic meaning of the key ``\textit{Format/Domain}" beforehand.
    Moreover, if we encounter a new input schema that involves new keys (e.g., ``\textit{extracted entities or commonsense knowledge}"), we can  add a new key-value pair in the input schema and learns the functionality of the new key indicator through training. 
    % In this way of adaptation, the model can still remember the semantic meaning of other components with learned key indicator representations.
    
    \textit{2)} \textit{Unified QA Systems:}
    We think further studies on unified QA systems could target on a better pre-training schema for general purpose QA, or optimizing the modeling strategy for the structural prompt to process more complex input, or output formats (e.g., adding extracted entities or retrieved knowledge).
    
    \textit{3)} \textit{Unification with Structural Prompt}: 
    The application of the structural prompt is not limited only on the QA task. 
    Intuitively, task inputs/outputs with various formats or components can also be organized with the structural prompt, like \textit{sentiment analysis} \cite{zhong-etal-2021-useradapter}, \textit{style transfer} \cite{li2022text}.
    In this way, we can integrate multiple tasks with carefully organized structural input, and improve the uniformity and expandability of the whole paradigm.
    
    % \item[\textit{3)}] \textit{Multi-task Learning with Structural Prompt:}
    % We can also potentially integrate the idea of structural prompt and masked language modeling to form a new paradigm of multi-task learning. 
    % Unlike traditional multi-task learning that designs specific prediction head for each sub-task and domain, we can achieve this by masking the value under different keys and ask the model to recover them. For example, we can mask the value for ``\textit{Sentiment}" key for sentiment analysis subtask, and mask the value for the ``\textit{Answer}" key for question answering. In this way, the model can better learn the common ability reside in different tasks with no needs for specifically designed prediction heads for multiple domains/tasks.

\section{Conclusion}
We introduce \modelname, a unified QA paradigm that adopts a single model for solving various QA tasks with the bridge of a structural prompt. 
Structural prompt simultaneously models the common ability required for various tasks and keeps the speciality of each task, through a structurally designed learnable input schema. 
We further conduct structural prompt-based pre-training, seeking to empower the model with general QA-centric ability and injects the semantic knowledge of the structural prompt into the pre-training model.
Experimental results on 11 QA benchmarks demonstrate that \modelname\ can significantly boost performance on all settings. 
Further analyses show that our method can better mitigate the catastrophic forgetting issue during continual learning, and our method can be adapted to a newly involved task more quickly, by taking the advantages of the structural prompt.
In the future, we hope our analysis could inspire more explorations on the unified QA methods, or the unification of distinct tasks with complex inputs modeling by the structural prompt.
We also hope structural prompt can be further utilized into the unification of more tasks with complex inputs. 
% , or can be used to improve the multi-task learning paradigm. 
\section{Acknowledgments}
Jian Yin is the corresponding author. Wanjun Zhong, Jiahai Wang and Jian Yin are supported by the Key-Area Research and Development Program of Guangdong Province (2020B0101100001).
\bibliography{anthology,custom}
\bibliographystyle{acl_natbib}

\clearpage

\appendix

\section{Implementation Details}\label{sec:app_implementation}
\subsection{Corpus Preparation}
In this part, we describe the details of corpus construction. 

The current pre-training corpus contains almost 4 million pre-training instances formulated with the structural prompt, including 1 million Multiple-choice QA instances, 2 million Extractive QA instances, and 2 million Abstractive QA instances.
When generating questions and answers, we take the {\em context} as the input, and the sequence ``{\em question \texttt{[SEP]} answer}" as the output.
In order to train the filtering model, we take the {\em context} and {\em question} as the inputs, and the {\em answer} as the output.
During the inference process of QA-pairs filtering, we take the context and the generated question as the model input of the QA model and set generated answer as the label. Then we compute the final soft score with the cross-entropy loss between the label and the answer generated by the QA model,
Next, we rerank all the generated QA-pairs according to the soft scores in an ascending order to select the most consistent QA-pairs as the pre-training instances.

Specifically, we employ AdamW as the optimizer for model training. We adopt \textit{T5-Large} as the model backbone and the seed datasets as supervisions for training both the question-answer pairs generation, and the filtering QA model.
We set learning rate as 1e-5, warmup step as 0, batch size as 2 per GPU, and training epochs as 10.

\subsection{Details on Pre-training and Task Adaptation.}
\paragraph{Pre-training.}
During pre-training, we jointly train the main model parameters with the representations of the \textit{special key indicators} and the \textit{task/format-specific soft prompts}.

Initially, we don't have any specific tasks during pre-training, so we take the three pre-training corpus (i.e., ``{\em MultiChoiceQA, Extractive QA, and Abstractive QA}" ) as the three initial tasks, and randomly initialize the task and format specific soft prompts. 

Specifically, we use \textit{T5-Base} as the model backbone, and set learning rate as 1e-4, batch size as 8 per GPU and gradient accumulation steps as 10. We adopt 8 V100 GPUs for pre-training. 
\paragraph{Fine-tuning.}
During fine-tuning, we need to initialize the task/format-specific soft prompts for a specific downstream task. 
If the task corresponds to a specific format participating in the pre-training stage, we use the corresponding soft prompts of this format type to initialize the soft prompts for the current tasks to transfer the learned knowledge. 
If the task corresponds to a new format, we can randomly initialize the task/format prompts.

Specifically, we use \textit{T5-Base} as the model backbone, and set learning rate as 1e-4, batch size as 2 per GPU, gradient accumulation steps as 2, and training epochs as 5. We adopt 8 V100 GPUs for fine-tuning.
\paragraph{Few-shot Learning.}
We adopt a similar way to initialize the task-specific soft prompts for few-shot learning. We use the standard setting which utilizes 32 randomly selected instances for few-shot learning. 
Specifically, we adopt \textit{T5-Base} as the model backbone, and set learning rate as 1e-5, batch size as 1 per GPU, gradient accumulation steps as 1, and training steps as 800 for few-shot learning. 
\paragraph{Zero-shot Learning}
Since zero-shot learning does not involve training stage, we just need to initialize the task-specific prompt for inference. Therefore, we initialize the task-specific prompt with the pre-trained task prompts of its corresponding format type. 

% Table generated by Excel2LaTeX from sheet 'latex'
\begin{table}[t]
  \centering
  \small
    % Table generated by Excel2LaTeX from sheet 'latex'
    \begin{tabular}{cl|c}
    \toprule
    Setting & \multicolumn{1}{c|}{Dataset} & BoolQ \\
    \midrule
    \multirow{2}[2]{*}{Full-Data} & T5    & 62.2 \\
          & Pro QA & 80.6 \\
    \midrule
    \multirow{2}[2]{*}{Few-Shot} & T5    & 0.0 \\
          & Pro QA & 55.4 \\
    \midrule
    \multirow{2}[2]{*}{Zero-Shot} & T5    & 0.0 \\
          & Pro QA & 62.1 \\
    \bottomrule
    \end{tabular}%
\caption{Result on two Yes/No QA tasks under full-data fine-tuning, few-shot learning, and zero-shot learning settings.}
\label{tab:yesno_result_appendix}%
\end{table}%

% Table generated by Excel2LaTeX from sheet 'latex'
\begin{table*}[t]
  \centering
  \resizebox{1.0\textwidth}{!}{
    \begin{tabular}{ccccc}
    \toprule
    Task A & \multicolumn{1}{p{10.165em}}{Task A Model \newline{}(Few-Shot Results)} & Task B & \multicolumn{1}{p{10.165em}}{Task B Model\newline{}(Evaluation on Task A)} & \multicolumn{1}{p{15em}}{Task B Model (w/ Task A Prompt)\newline{}(Evaluation on Task A)} \\
    \midrule
    \multirow{6}[2]{*}{EX/NewsQA (EM)} & \multirow{6}[2]{*}{42.1} & EX/Quoref (EM) & 33.7  & 35.3 \\
          &       & AB/DROP (F1) & 31.3  & 33.7 \\
          &       & AB/NQOpen (F1) & 27.6  & 34.2 \\
          &       & MC/DREAM (Acc) & 35.2  & 37.9 \\
          &       & MC/MCTest (Acc) & 34.5  & 36.8 \\
          &       & MC/OBQA (Acc) & 34.4  & 35.9 \\
    \midrule
    \multirow{6}[2]{*}{EX/Quoref (EM)} & \multirow{6}[2]{*}{52.2} & EX/NewsQA (EM) & 41.1  & 42.8 \\
          &       & AB/DROP (F1) & 43.0  & 51.4 \\
          &       & AB/NQOpen (F1) & 38.0  & 51.0 \\
          &       & MC/DREAM (Acc) & 47.2  & 51.3 \\
          &       & MC/MCTest (Acc) & 47.8  & 52.0 \\
          &       & MC/OBQA (Acc) & 48.5  & 51.6 \\
    \midrule
    \multirow{3}[2]{*}{AB/NQOpen (F1)} & \multirow{3}[2]{*}{36.0} & EX/NewsQA (EM) & 32.8  & 33.9 \\
          &       & EX/Quoref (EM) & 36.2  & 35.4 \\
          &       & AB/DROP (F1) & 34.3  & 35.8 \\
    \midrule
    \multirow{6}[2]{*}{AB/DROP (F1)} & \multirow{6}[2]{*}{27.1} & EX/NewsQA (EM) & 22.8  & 26.3 \\
          &       & EX/Quoref (EM) & 24.0  & 26.6 \\
          &       & AB/NQOpen (F1) & 25.0  & 25.6 \\
          &       & MC/DREAM (Acc) & 25.1  & 26.2 \\
          &       & MC/MCTest (Acc) & 25.4  & 26.7 \\
          &       & MC/OBQA (Acc) & 25.6  & 26.7 \\
    \midrule
    \multirow{4}[2]{*}{MC/MCTest (Acc)} & \multirow{4}[2]{*}{84.1} & EX/NewsQA (EM) & 81.6  & 83.1 \\
          &       & AB/DROP (F1) & 82.8  & 83.8 \\
          &       & MC/DREAM (Acc) & 83.2  & 83.8 \\
          &       & MC/OBQA (Acc) & 82.2  & 82.5 \\
    \midrule
    \multirow{4}[2]{*}{MC/DREAM (Acc)} & \multirow{4}[2]{*}{66.0} & EX/NewsQA (EM) & 64.0  & 65.4 \\
          &       & AB/DROP (F1) & 63.5  & 65.7 \\
          &       & MC/MCTest (Acc) & 65.5  & 65.8 \\
          &       & MC/OBQA (Acc) & 65.2  & 65.5 \\
    \midrule
    \multirow{4}[2]{*}{MC/OBQA (Acc)} & \multirow{4}[2]{*}{44.8} & EX/NewsQA (EM) & 41.4  & 45.2 \\
          &       & AB/DROP (F1) & 40.6  & 44.2 \\
          &       & MC/MCTest (Acc) & 44.8  & 45.8 \\
          &       & MC/DREAM (Acc) & 44.2  & 46.8 \\
    \bottomrule
    \end{tabular}%
    }
\caption{
Full results on continual learning.
For each task, we provide the task format (EX, AB, MC) and its evaluation metrics (EM, F1, Acc).
EX: Extractive QA; AB: Abstractive QA; MC: Multiple-Choice QA.
}
\label{tab:continual_appendix}%
\end{table*}%

\section{Results on Yes/No Pre-training}\label{sec:app_results_yesno}

During our pilot study, we take the BoolQ \cite{clark-etal-2019-boolq} as the seed dataset to construct a large-scale pre-training corpus, and test the full-data, few-shot, zero-shot on top of the pre-trained \modelname.
We also take the naturally-perturbed version of this dataset BoolQ-NP \cite{khashabi-etal-2020-bang} into account for evaluation.
Results are shown in Table~\ref{tab:yesno_result_appendix}.
We find that the \modelname\ significantly outperforms T5 baseline on all settings.
% Additionally, \modelname\ achieves 100\% accuracy for BoolQ-NP on all evaluation settings.
Note that we take a strict evaluation towards the model's output.
In other words, if the output is not any format of ``yes'', ``no'', ``true'', ``false'', that prediction will be classified as wrong.

\section{Details on Continual Learning}\label{sec:app_results_continual}
Table~\ref{tab:continual_appendix} provides the full results for the continual learning experiment.
The model is firstly trained on task A under few-shot setting, and then fine-tuned on task B.
Afterwards, we evaluate the trained ``Task B Model'' and ``Task B Model (w/ Task A Prompt)'' on task A to test its continual learning capability.
Detailed results on every Task A/Task B combination (33 reported in total) are shown in Table~\ref{tab:continual_appendix}.
% We report results on 33 combinations of datasets. 
% We stop the continual learning experiments on task B because we want to more directly observe the influence of task adaptation between tasks with different formats. 
% We select the combinations of two tasks, because it is a direct way to evaluate the influence of task adaptation between different formats, besides only evaluating the performance drop on continual learning.
Note that we consider two tasks in continual learning because we also want to investigate the task adaptation to-and-fro the same format (e.g., MC $\rightarrow$ MC) or different formats (e.g., AB $\rightarrow$ EX).
The results shed light on how could we arrange the order of training on tasks to achieve the best overall performance when a bunch of tasks arrive.

% \section{Additional Results for Ablation Study}
% We 

\end{document}